\documentclass[10pt,twocolumn,letterpaper]{article}

\usepackage[pagenumbers]{wacv} % To force page numbers, e.g. for an arXiv version
\usepackage{graphicx}
\usepackage{amsmath}
\usepackage{amssymb}
\usepackage{booktabs}
\usepackage{subcaption}
\usepackage{caption}
\usepackage{bm}
\usepackage{xcolor}

\usepackage{array}    % For table formatting
\usepackage{multirow} % For multi-row cells
\usepackage{colortbl} % Allows coloring of individual cells
\usepackage{color} %using the color package, not xcolor

\usepackage[pagebackref,breaklinks,colorlinks]{hyperref}

\usepackage[capitalize]{cleveref}
\crefname{section}{Sec.}{Secs.}
\Crefname{section}{Section}{Sections}
\Crefname{table}{Table}{Tables}
\crefname{table}{Tab.}{Tabs.}

\def\wacvPaperID{577} % *** Enter the WACV Paper ID here
\def\confName{WACV}
\def\confYear{2025}

\newcommand{\zt}[1]{{\color{black}#1}}

\begin{document}

%%%%%%%%% TITLE - PLEASE UPDATE
\title{Textured-GS: Gaussian Splatting with Spatially Defined Color and Opacity}

\author{Zhentao Huang\\
University of Guelph\\
School of Computer Science\\
{\tt\small zhentao@uoguelph.ca}
% For a paper whose authors are all at the same institution,
% omit the following lines up until the closing ``}''.
% Additional authors and addresses can be added with ``\and'',
% just like the second author.
% To save space, use either the email address or home page, not both
\and
Minglun Gong\\
University of Guelph\\
School of Computer Science\\
{\tt\small minglun@uoguelph.ca}
}
\maketitle
\begin{figure*}[t!]
    \centering
    \begin{subfigure}[b]{0.3\textwidth}
        \centering
        \includegraphics[width=1.0\textwidth]{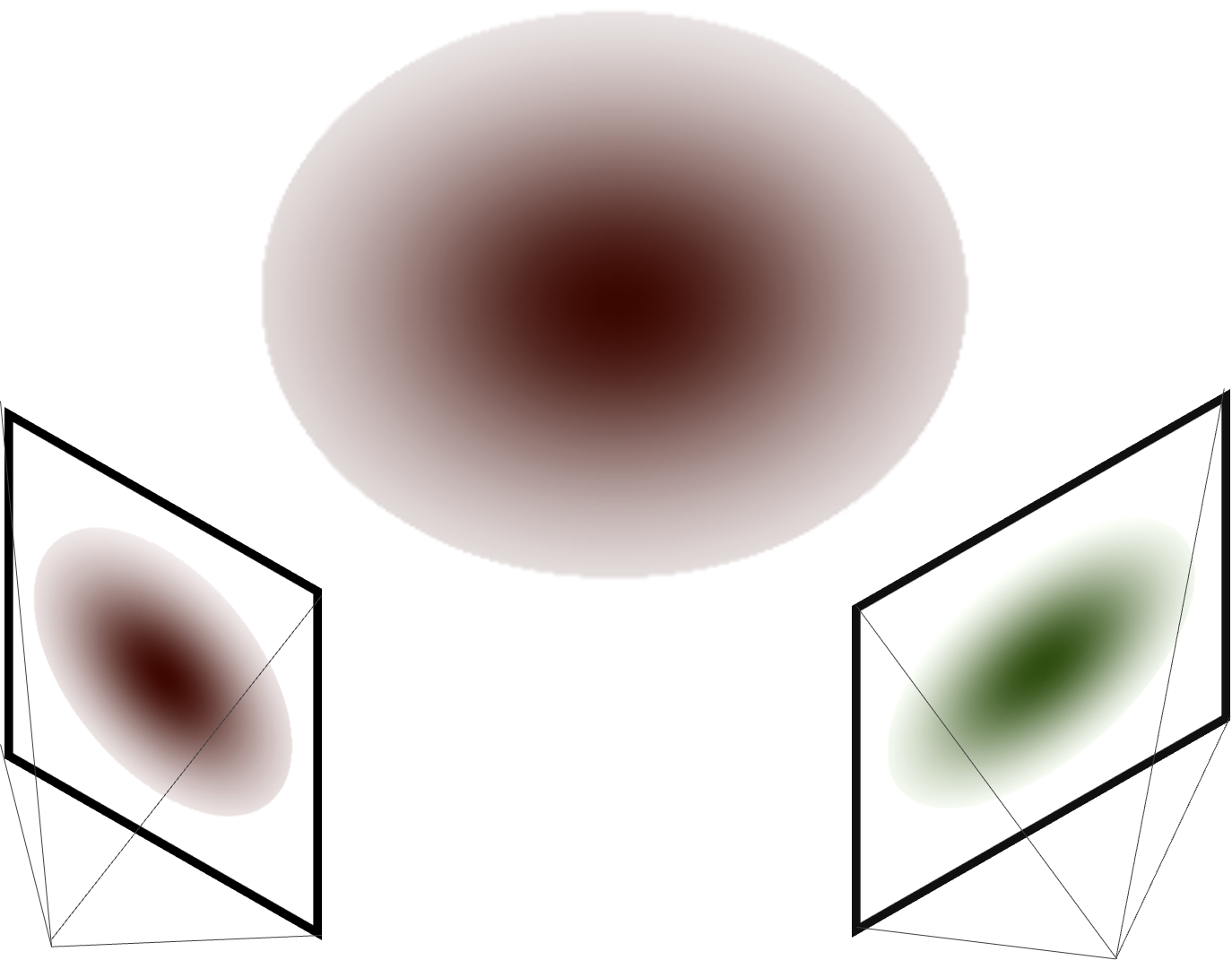}
        \caption{3D Gaussian}
        \label{fig:subfig1}
    \end{subfigure}
    \hfill
    \begin{subfigure}[b]{0.3\textwidth}
        \centering
        \includegraphics[width=1.0\textwidth]{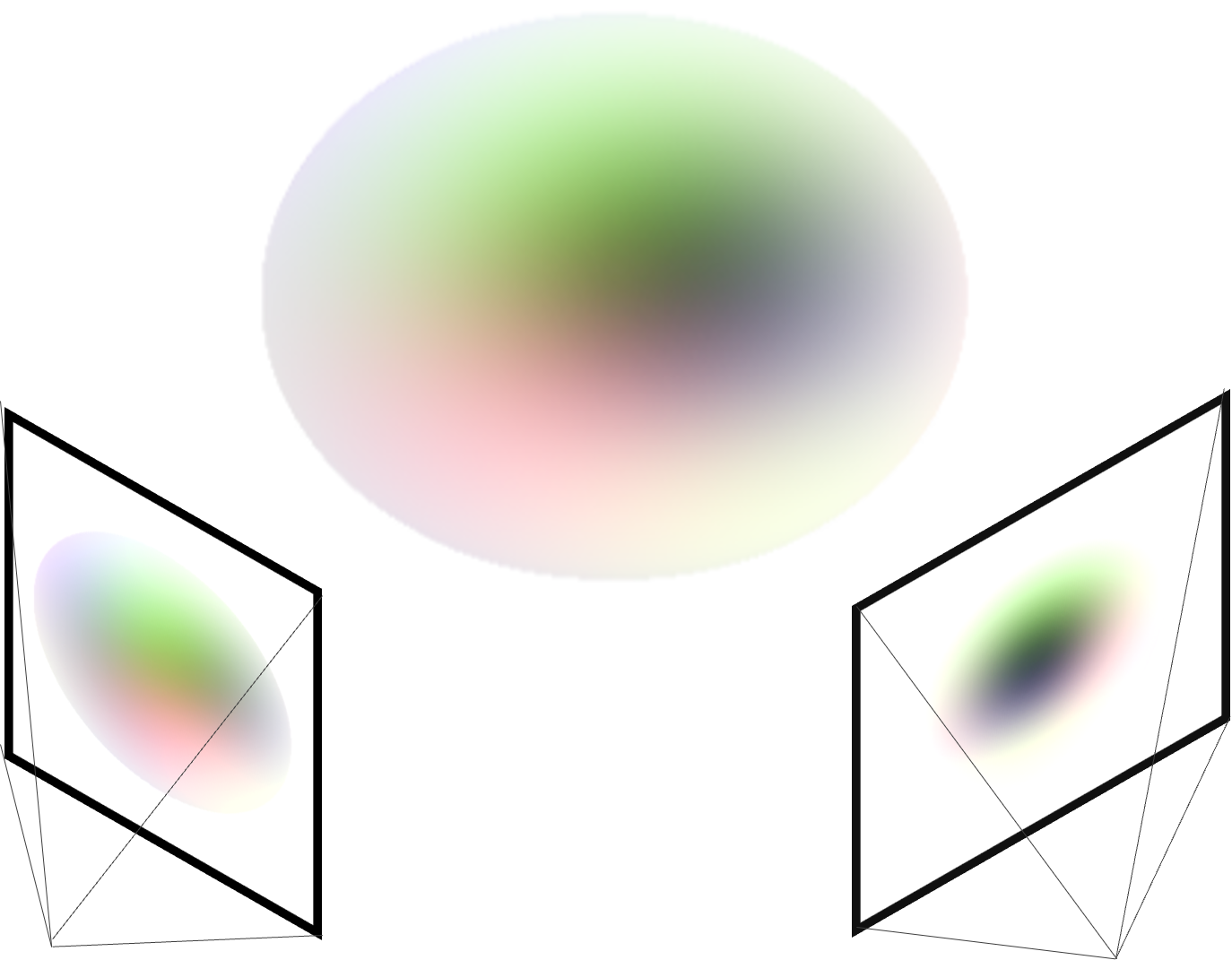}
        \caption{\zt{Gaussian with Color Texture}}
        \label{fig:subfig2}
    \end{subfigure}
    \hfill
    \begin{subfigure}[b]{0.3\textwidth}
        \centering
        \includegraphics[width=1.0\textwidth]{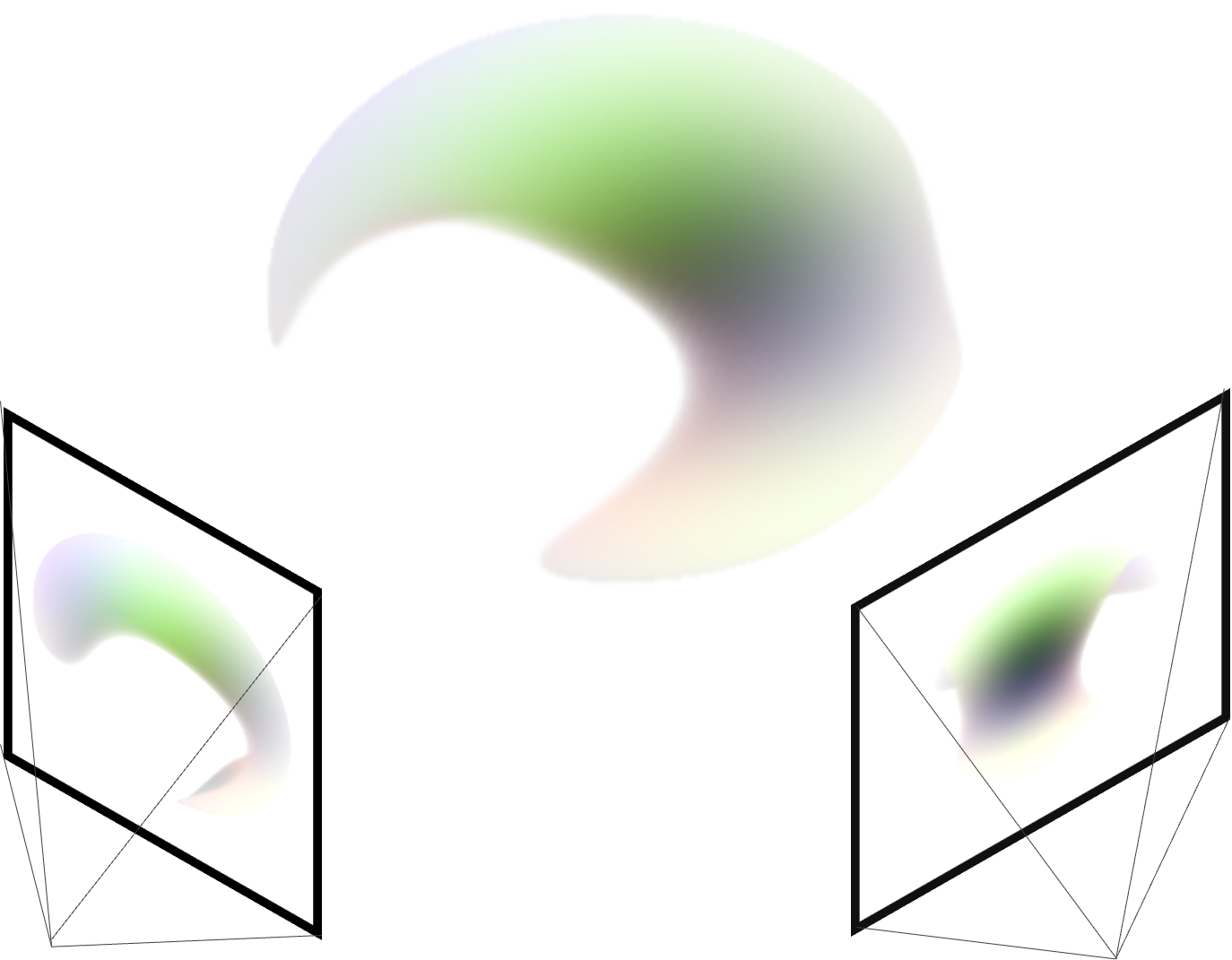}
        \caption{\zt{Gaussian with Color \& Opacity Texture}}
        \label{fig:subfig3}
    \end{subfigure}
    \caption{Comparison between view-dependent 3D Gaussian and our textured Gaussian: The original 3DGS \cite{kerbl20233d} (a) assigns a single opacity value to each Gaussian and maintains a fixed color per view. In contrast, our Textured-GS (b) enables color variation across the Gaussian ellipsoidal surface. With the addition of an opacity channel (c), it can also represent non-ellipsoidal shapes.}
    \label{fig:spatial}
\end{figure*}

%%%%%%%%% ABSTRACT
\begin{abstract}
    In this paper, we introduce Textured-GS, an innovative method for rendering Gaussian splatting that incorporates spatially defined color and opacity variations using Spherical Harmonics (SH). This approach enables each Gaussian to exhibit a richer representation by accommodating varying colors and opacities across its surface, significantly enhancing rendering quality compared to traditional methods. To demonstrate the merits of our approach, we adapted the Mini-Splatting architecture to integrate textured Gaussians without increasing the number of Gaussians. Our experiments across multiple real-world datasets show that Textured-GS consistently outperforms both the baseline Mini-Splatting and standard 3DGS in terms of visual fidelity. These results highlight the potential of Textured-GS to advance Gaussian-based rendering technologies, promising more efficient and high-quality scene reconstructions. Our implementation is available at \url{https://github.com/ZhentaoHuang/Textured-GS}.
\end{abstract}

%%%%%%%%% BODY TEXT
\section{Introduction}
\label{sec:intro}

Novel view synthesis is an active topic in Computer Vision and Graphics, focusing on generating accurate and realistic views from sparse input images with known camera parameters. Neural radiance fields (NeRF) \cite{mildenhall2021nerf} introduce a method to model a 3D scene by optimizing a continuous volumetric scene function through a fully connected deep network. This approach leverages implicit neural representations, extracting features like colors and geometry from a 3D scene by querying simple multi-layer perceptron networks with 5D inputs (spatial locations and camera views). The impressive realism achieved in free-viewpoint rendering by this learned mapping has sparked a surge of follow-up methods aimed at improving quality and speed, often incorporating regularization strategies. Notable examples include Mip-NeRF360 \cite{barron2022mip} and Zip-NeRF \cite{barron2023zip}, which excel in rendering quality, although they still require substantial training and rendering times. InstantNGP \cite{muller2022instant} enhances efficiency with multiresolution hash encoding and a streamlined network architecture, reducing both the training duration and the computational demands during inference. However, there are scenarios where NeRF's rendering capability might not fully capture detailed scene elements, particularly in complex light interactions and reflections. Furthermore, NeRF-based methods often struggle to accurately represent empty spaces.

More recently, 3D Gaussian Splatting (3DGS) \cite{kerbl20233d} has been introduced as a method for real-time rendering that significantly accelerates both the rendering process and scene optimization through a set of optimized 3D Gaussians, efficiently rasterizable on modern GPUs. In addition to its speed, 3DGS achieves a rendering quality that matches or even surpasses the leading NeRF implementations. Scenes are represented by millions of 3D Gaussians with specified position, rotation, scale, opacity, and color parameters, requiring significant storage and memory, rendering it impractical on devices with limited video memory like smartphones or head-mounted displays.

To address storage issues, Niedermayr et al. \cite{niedermayr2024compressed} have shown that Spherical Harmonic (SH) coefficients can be considerably redundant, proposing their compression into compact codebooks. Fang and Wang \cite{fang2024mini} point out the inefficient spatial distribution of Gaussian representations as a key bottleneck in rendering performance. However, a Gaussian represented with SH coefficients exhibits only a single color from a specific viewing angle, limiting its representational capability.

Our objective is to enhance the textural representation of individual Gaussians to model color variations locally, even from a single viewing direction. By maintaining the existing SH framework and parameter allocation, we modify the parameterization scheme. This adjustment allows each 3D Gaussian to display different colors at various viewing angles and across different areas of the Gaussian ellipsoidal surface when viewed from the same angle. Additionally, we integrate an opacity channel into the SH framework, enabling the modeling of opacity variations along the Gaussian surface, thereby enriching visual complexity and increasing the realism of rendered scenes.

In summary, our work presents several key contributions:
\begin{itemize}
\item We introduce a novel method that textures Gaussian ellipsoidal surfaces using SH, increasing the representational power of individual Gaussians without adding extra parameters.
\item We extended this technique to model opacity variations on Gaussian ellipsoidal surfaces, allowing for deviations from the standard ellipsoidal shape. This includes creating ``unidirectional Gaussians" that can be opaque on one side and transparent on the other.
\item We validated our approach by applying it to pre-optimized 3D Gaussian scenes using mini-splatting \cite{fang2024mini}. Our method achieves comparable rendering quality on established datasets while requiring significantly fewer number of Gaussians compared to standard 3D Gaussian Splatting (3DGS) \cite{kerbl20233d}.
\end{itemize}

% Recent progress in 3D Gaussian Splatting (3DGS) \cite{kerbl20233d} has demonstrated its considerable promise in various applications, including immersive rendering and 3D reconstruction, due to its ability to provide real-time and high-quality rendering. 

%------------------------------------------------------------------------
\section{Related Work}
%TODO: check grammar
\label{sec:related}
\subsection{Novel View Synthesis}
Novel view synthesis (NVS) has emerged as a pivotal technique in computer graphics and vision, facilitating the generation of new perspectives from a sparse set of images through advanced modeling of 3D scenes. Among the significant advancements in this domain, Neural Radiance Fields (NeRF) and 3D Gaussian Splatting (3DGS) represent transformative approaches that have reshaped our capabilities for rendering complex, photorealistic scenes efficiently. 

Introduced by Mildenhall et al. \cite{mildenhall2021nerf}, Neural Radiance Fields (NeRF) leverage a fully connected neural network to continuously model a volumetric scene function, enabling precise capture of color and density throughout 3D space. NeRF generates new perspectives with notable detail and realism by processing these predictions through a differentiable rendering framework along the viewer's line of sight. Despite its impressive output quality, the main drawback of NeRF is its computational intensity and slow rendering times, which have led to numerous subsequent studies aimed at overcoming these challenges \cite{garbin2021fastnerf, chen2023mobilenerf, muller2022instant, fridovich2022plenoxels, lindell2021autoint, reiser2021kilonerf, kosiorek2021nerf}.

Key advancements in this area include Instant-NGP \cite{muller2022instant}, which employs a neural hash grid to markedly decrease computational load while sustaining output quality. Plenoxels \cite{fridovich2022plenoxels} utilize a sparse, learnable grid that eschews traditional neural networks, significantly expediting both training and inference. Mip-NeRF 360 \cite{barron2022mip} focuses on improving renderings within 360-degree environments, tackling the challenges associated with view variability and inconsistent lighting. Furthermore, Zip-NeRF \cite{barron2023zip} innovates in network compression methods to enhance rendering fidelity without sacrificing the quality of the produced images.

\subsection{3D Gaussian Splatting Compression}
Developed by Kerbl et al. as an efficient alternative to traditional volume rendering techniques, 3DGS \cite{kerbl20233d} represents scene through a sparse set of 3D Gaussians. Each Gaussian in this representation is characterized by its position, color, and covariance matrix. 3DGS stands out due to its differentiable rendering process, allowing it to be optimized directly from photometric observations. This technique not only speeds up the rendering process but also enhances the adaptability of the representation, making it particularly effective for real-time applications such as VR and AR where rapid rendering is crucial. Although the 3D Gaussian Splatting achieves real-time rendering, there is improvement space in terms of lower computational requirements and better point distribution \cite{wu2024recent}. 

Various 3DGS compression techniques employ vector quantization to cluster multi-dimensional data into a finite set of representations \cite{lee2024compact, navaneet2023compact3d, niedermayr2024compressed, girish2023eagles, fan2023lightgaussian}. Specifically, Niedermayr et al. \cite{niedermayr2024compressed} utilizes vector clustering to compactly encode color and geometric attributes into two codebooks to reduce redundancy. Similarly, the EAGLES \cite{girish2023eagles} applies quantization across all attributes of each Gaussian and show that the quantization of opacity leads to fewer floaters or visual artifacts. However, these techniques often do not address the suboptimal distribution of Gaussians, which tends to result in local minima following compression. \zt{We believe that similar compression techniques can also be applied on Textured-GS in the future work, as there are four channels of SH coefficients to be compressed.}

Rather than simply compressing existing 3D Gaussian Splatting systems, several initiatives aim to improve Gaussian distribution \cite{fang2024mini, fan2023lightgaussian, morgenstern2023compact, jo2024identifying}. LightGaussian \cite{fan2023lightgaussian} reduces the number of Gaussians by pruning those with lower importance scores and employs an octree-based method for compressing positions. Jo et al. \cite{jo2024identifying} developed a strategy to both compress 3DGS and enhance computational efficiency by eliminating non-essential Gaussians. Meanwhile, Mini-Splatting \cite{fang2024mini} enhances the rendering process by utilizing more effective Gaussian splats. This method samples Gaussians based on importance score instead of pruning them to avoid artifacts. \zt{It reduced the scenario where multiple Gaussians are stacked to form one color, which aligns with our objective of applying textures to individual Gaussians. Therefore, it has been selected as our baseline method.}

% 3DGS-Avatar \cite{qian20243dgs} introduces an approach that creates animatable human avatars from monocular videos using 3D Gaussian
% Splatting (3DGS).

\begin{table*}[htbp]
\centering

\resizebox{\textwidth}{!}{% Resize table to fit within the textwidth
\begin{tabular}{|c|cccc|cccc|cccc|}
\hline
\multirow{2}{*}{\begin{tabular}[c]{@{}c@{}}\textbf{Method} \vline \ \textbf{Metric}\end{tabular}} & \multicolumn{4}{c|}{\textbf{Mip-NeRF 360}} & \multicolumn{4}{c|}{\textbf{Tanks\&Temples}} & \multicolumn{4}{c|}{\textbf{Deep Blending}} \\ \cline{2-13}
 & \textbf{SSIM} $\uparrow$ & \textbf{PSNR} $\uparrow$ & \textbf{LPIPS} $\downarrow$ & \textbf{Num} & \textbf{SSIM} $\uparrow$ & \textbf{PSNR} $\uparrow$ & \textbf{LPIPS} $\downarrow$ & \textbf{Num} & \textbf{SSIM} $\uparrow$ & \textbf{PSNR} $\uparrow$ & \textbf{LPIPS} $\downarrow$ & \textbf{Num} \\ \hline
Plenoxels \cite{fridovich2022plenoxels} & 0.626 & 23.08 & 0.463 & - & 0.719 & 21.08 & 0.379 & - & 0.795 & 23.06 & 0.510 & - \\
INGP-Big \cite{muller2022instant} & 0.699 & 25.59 & 0.331 & - & 0.745 & 21.92 & 0.305 & - & 0.817 & 24.96 & 0.390 & - \\
mip-NeRF 360 \cite{barron2022mip} & 0.792 & \colorbox{yellow}{27.69} & 0.237 & - & 0.759 & 22.22 & 0.257 & - & 0.901 & 29.40 & \colorbox{yellow}{0.245} & - \\
Zip-NeRF \cite{barron2023zip} & \colorbox{orange}{0.828} & \colorbox{red}{28.54} & \colorbox{orange}{0.189} & - & - & - & - & - & - & - & - & - \\ \hline
3DGS \cite{kerbl20233d} & 0.815 & 27.21 & 0.214 & \textbf{3.36} & 0.841 & 23.14 & \colorbox{orange}{0.183} & \textbf{1.78} & 0.903 & 29.41 & \colorbox{orange}{0.243} & \textbf{2.98} \\ 

Mip-Splatting \cite{yu2024mip} & \colorbox{yellow}{0.827} & \colorbox{orange}{27.79} & \colorbox{yellow}{0.203} &  \textbf{3.97} & - & - & - & - & - & - & - & - \\

3DGS-MCMC \cite{kheradmand20243d} & - & - & - & - & \colorbox{red}{0.860} & \colorbox{red}{24.29} & \colorbox{yellow}{0.190} & \textbf{1.78} & 0.890 & 29.67 & 0.320 & \textbf{2.98} \\

Mini-Splatting-D \cite{fang2024mini} & \colorbox{red}{0.831} & 27.51 & \colorbox{red}{0.176} & \textbf{4.69} & \colorbox{orange}{0.853} & 23.23 & \colorbox{red}{0.140} & \textbf{4.28} & \colorbox{yellow}{0.906} & \colorbox{yellow}{29.88} & \colorbox{red}{0.211} & \textbf{4.63} \\

Mini-Splatting (30K) \cite{fang2024mini} & 0.822 & 27.34 & 0.217 & \textbf{0.49} & 0.835 & 23.18 & 0.202 & \textbf{0.20} & \colorbox{orange}{0.908} & \colorbox{orange}{29.98} & 0.253 & \textbf{0.35} \\ 

Mini-Splatting (44K) \cite{fang2024mini} & 0.822 & 27.34 & 0.213 & \textbf{0.49} & 0.840 & \colorbox{yellow}{23.34} & 0.203 & \textbf{0.20} & \colorbox{yellow}{0.906} & 29.80 & 0.250 & \textbf{0.35} \\ 
Textured-GS & 0.825 & 27.64 & 0.209 & \textbf{0.49} & \colorbox{yellow}{0.843} & \colorbox{orange}{23.49} & 0.191 & \textbf{0.20} & \colorbox{red}{0.909} & \colorbox{red}{30.02} & 0.248 & \textbf{0.35} \\ 
\hline
\end{tabular}
}
\caption{Quantitative evaluation of our Textured-GS and previous works in three commonly-used metrics: SSIM, PSNR and LPIPS. Num represent number of Gaussians (million).}
\label{tab:All}
\end{table*}

\section{3D Gaussian Splatting Preliminaries}

% \textbf{Standard 3D Gaussian Splatting} 
3D Gaussian Splatting (3DGS) leverages the principles of Elliptical Weighted Average (EWA) volume splatting \cite{zwicker2001ewa} as the foundational technique for the efficient computation of 3D Gaussian kernel projections onto a 2D image plane. Building upon this, the method employs differentiable Gaussian splatting \cite{kerbl20233d} to further refine the rendering process. This approach enables dynamic optimization of both the amount and the specific parameters of Gaussian kernels used for scene modeling.

The optimized scene consists of millions of 3D Gaussians, each defined by a set of parameters: $\bm{x} \in \mathbb{R}^3$ for center position, $\bm{q} \in \mathbb{R}^4$ for the quaternion that represents the rotation, $\bm{s} \in \mathbb{R}^3$ for the scale of the Gaussian in each dimension, $\alpha \in [0,1]$ for the opacity, and 16 spherical harmonics (SH) coefficients for view dependent coloring. The covariance matrix of the Gaussian is calculated as follows:
\begin{equation}
    \bm{\Sigma} = R S S^T R^T
\end{equation}
where $R$ is the rotation matrix derived from the quaternion $q$ and $S$ is a diagonal scaling matrix constructed from the scale vector $s$. This formulation allows for the independent optimization of rotation and scaling, facilitating more flexible and precise control over the Gaussian's appearance in the rendered scene.

In the rendering process, the projection of a 3D Gaussian onto the 2D image plane is mathematically captured by transforming its covariance matrix:
\begin{equation}
    \bm{\Sigma'} = J W \Sigma \mathit{W}^T \mathit{J}^T,
\end{equation}
where $J$ is the Jacobian of the affine approximation of the projective transformation, and $W$ is the viewing transformation matrix that translates and rotates the 3D Gaussian from world coordinates into camera coordinates. This allows to evaluate the 2D color and opacity footprint of each projected Gaussian. The final pixel color $C$ is computed by blending $N$ 2D Gaussians that contribute to this pixel, sorted in order of their depth:
\begin{eqnarray}
C &=& \sum_{i=1}^N w_i \cdot c_i, \label{eq:C1-1} \\ 
w_i &=& T_i \cdot \alpha_i \cdot G_i^{2D}(\bm{x}), \label{eq:C1-2} \\
c_i &=&  \sum_{l=0}^L \sum_{m=-l}^l \bm{\gamma}_{lm}(i) Y_{lm}(\bm{v_i}), \label{eq:C1-3} 
\end{eqnarray}
where the contribution of the $i^{th}$ Gaussian ($w_i$) is determined by its opacity $\alpha_i$, its projected distribution $G_i^{2D}(x)$, and the accumulated opacity of Gaussians in front of it, represented as $T_i = \prod_{j=1}^{i-1} (1 - w_j)$. The color of the Gaussian ($c_i$) is computed using the SH function, where $L$ represents the maximum degree of SH coefficients and $Y_{lm}$ denotes the pre-defined SH basis functions \cite{ramamoorthi2001efficient}.

Please note that, in the described formulation, each 3D Gaussian retains a fixed opacity and shows a single color when viewed from a specific direction.

\section{Methodology}

\textbf{Textured-GS} 3D Gaussian Splatting (3DGS) \cite{kerbl20233d} utilizes Spherical Harmonics (SH) to model view-dependent shading effects. The SH basis functions are evaluated based on the view vector direction, defined by the vector from the camera center to the Gaussian center. These evaluations are then multiplied by the SH coefficients. The summed products of the SH basis functions and coefficients determine the view-dependent color for a given Gaussian, reflecting both the environmental lighting and dynamic changes based on the viewer’s position relative to the object's surface. However, each Gaussian's color remains fixed under a specific view direction, regardless of the Gaussian's size, limiting the model's ability to represent high-frequency color variations across the Gaussian surface. Furthermore, in scenarios consisting solely of Lambertian surfaces, the additional SH parameters become redundant, as there are no view-dependent shading effects.

To address the limitations of traditional 3DGS, we introduce a novel approach utilizing SH coefficients to texture Gaussian ellipsoidal surfaces. Our method enables a single Gaussian to model local color and opacity variations, significantly enhancing its representational power and improving the overall quality of scene rendering with the same number of Gaussians.

\begin{figure}
    \centering
    \includegraphics[width=1\linewidth]{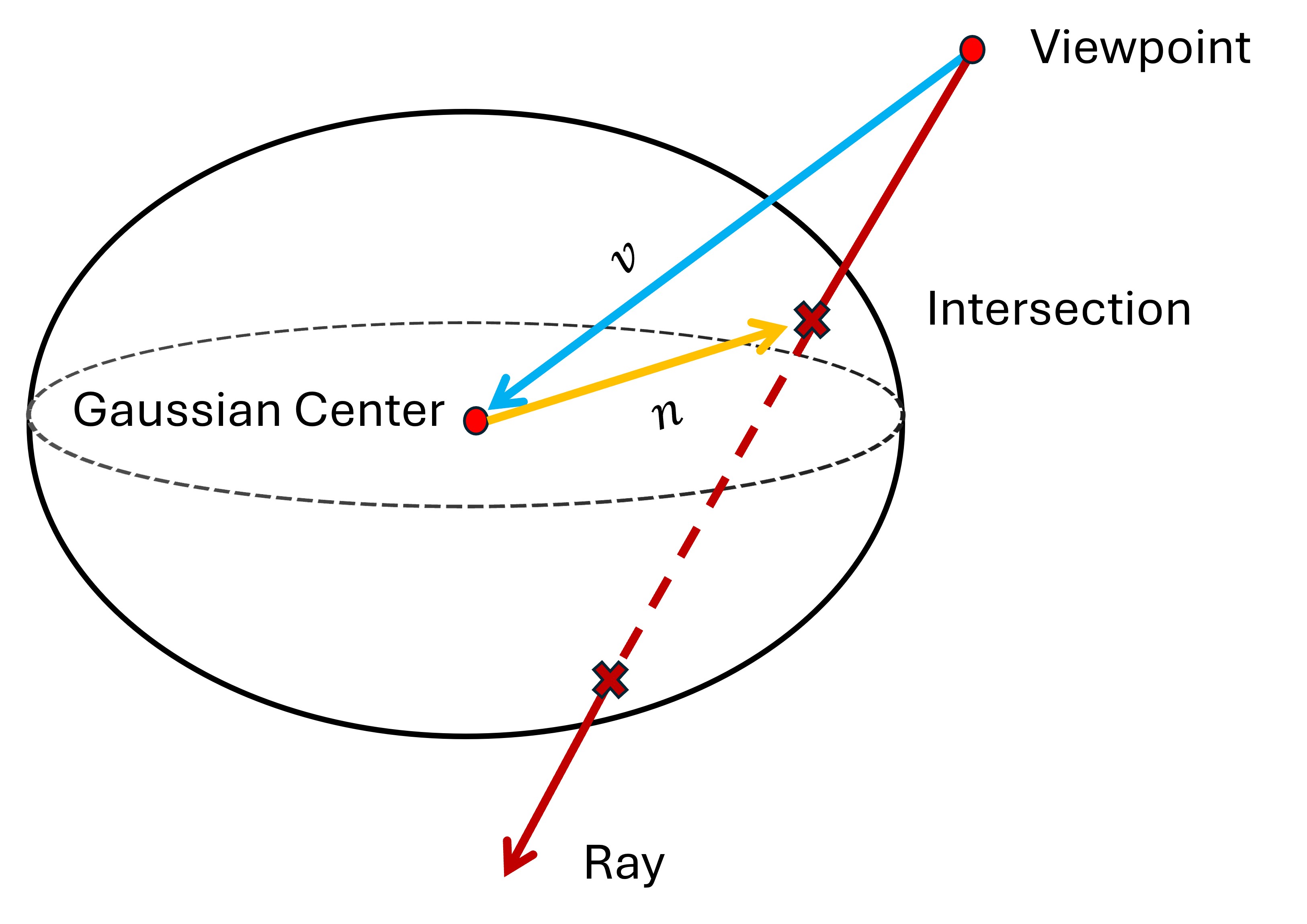}
    \caption{Illustration of the vectors used for SH calculation: The standard 3DGS employs the viewing vector  $\bm{v}$ (blue), which remains constant for all pixels under a given view. In contrast, our proposed Textured-GS uses the parameterization vector $\bm{n}$ (orange), which varies across the Gaussian surface. Please note that $\bm{n}$ is normalized by the scale of the Gaussian of each dimension.}
    \label{fig:intersection}
\end{figure}

Specifically, we calculate the intersection between the viewing ray and the Gaussian's ellipsoidal surface, as demonstrated in Figure \ref{fig:intersection}. This intersection point is then transformed into the Gaussian's local coordinate system and normalized according to the scale of the Gaussian in each dimension, enabling distinct parameterization of different areas on the Gaussian surface. That is, the normalized vector $n$, instead of viewing ray $v$, is used to evaluate SH basis functions, which allows the Gaussian to form a texture on the surface of the ellipsoid, as illustrated in Figure \ref{fig:spatial}.
The complexity of the texture depends on the degree level of SH coefficients. In this paper, we select level three, which uses the same number of parameters as in 3DGS \cite{kerbl20233d}.

The new formulation further enhances our ability to model opacity variations across the Gaussian surface by incorporating an opacity channel into each SH coefficient. As demonstrated in Figure \ref{fig:spatial}, enabling variable opacity across a Gaussian's surface can effectively alter its perceived shape. 

With both color and opacity textured, we replaces Equation (\ref{eq:C1-3}) with the following:
\begin{eqnarray}
c_i &=&  \sum_{l=0}^L \sum_{m=-l}^l \bm{\gamma}_{lm}(i) Y_{lm}(\bm{n}), \\
\alpha_i &=&  \sum_{l=0}^L \sum_{m=-l}^l \bm{\alpha}_{lm}(i) Y_{lm}(\bm{n}),
\label{eq:C2}
\end{eqnarray}

Now, we present our formulation of the view vector evaluated by the SH coefficients. We model the Gaussian as an ellipsoid in 3D space:\zt{
\begin{equation}
    g(x,y,z) = \frac{x^2}{s_x^2} + \frac{y^2}{s_y^2} + \frac{z^2}{z_x^2} = 1,
    \label{eq:ellipsoid}
\end{equation}
where $s_x,s_y,s_z$ represent the scale of the Gaussian in each dimension, respectively. Given an input pixel-ray $\bm{r}(t) = \bm{o} + t\bm{d}$, where $\bm{o}$ and $\bm{d}$ are the ray's origin and the direction transformed into the ellipsoid's local coordinate system, we calculate the position of the two intersection points. Substituting the parametric ray equation into the ellipsoid equation results in:
\[
\frac{(o_x + t d_x)^2}{s_x^2} + \frac{(o_y + t d_y)^2}{s_y^2} + \frac{(o_z + t d_z)^2}{s_z^2} = 1
\]

After expanding and simplifying, we obtain a quadratic equation in terms of $t$:

The quadratic equation then becomes:
\[
At^2 + Bt + C = 0,
\]

\begin{align}
\text{where } A &= \frac{d_x^2}{s_x^2} + \frac{d_y^2}{s_y^2} + \frac{d_z^2}{s_z^2}, \nonumber \\
B &= 2(\frac{o_x d_x}{s_x^2} + \frac{o_y d_y}{s_y^2} + \frac{o_z d_z}{s_z^2}), \notag \\
C &= \frac{o_x^2}{s_x^2} + \frac{o_y^2}{s_y^2} + \frac{o_z^2}{s_z^2} - 1 \notag.
\end{align}

The $t$ can be solved using the quadratic formula:

\[
t = \frac{-B \pm \sqrt{B^2-4AC}}{2A}
\]
We select the intersection point with lower $t$ which means it is closer to the ray's origin, and normalized each dimension based on the scale $s$. 
}

\begin{figure*}[htbp]
    \centering
    \begin{subfigure}[b]{\textwidth}  % Adjust the width to fit your specific needs
        \centering
        \includegraphics[width=\textwidth]{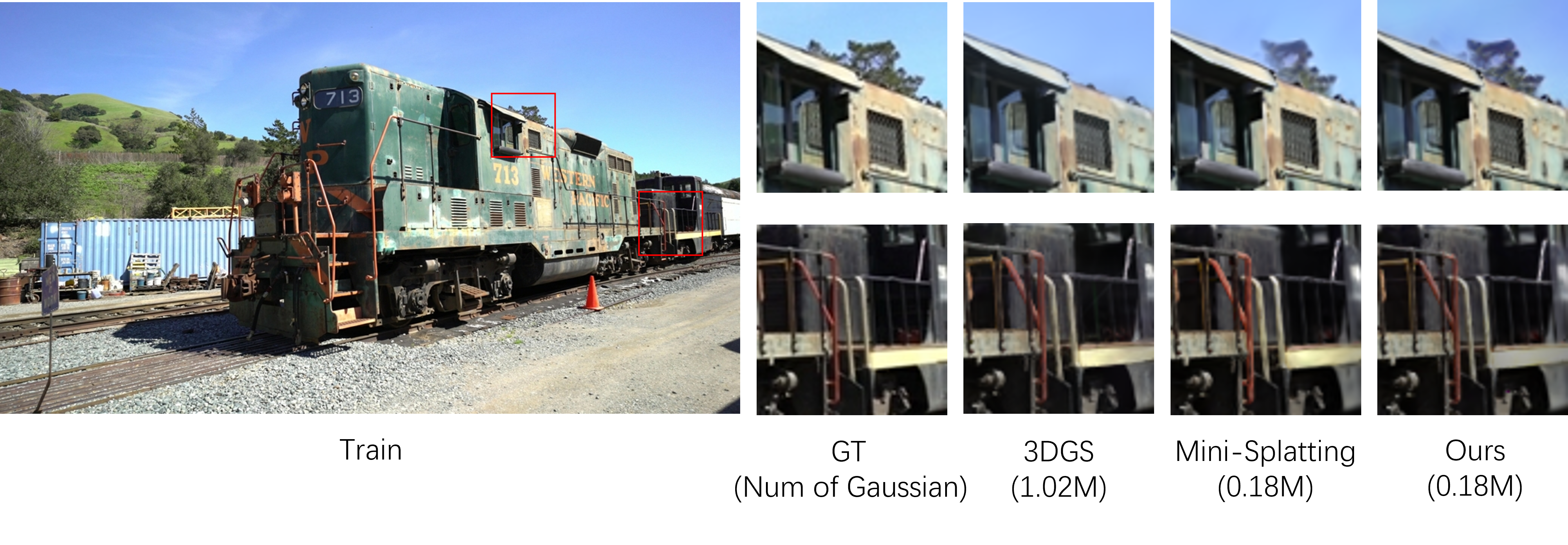}
        % \caption{First subfigure}
        \label{fig:sub1}
    \end{subfigure}

    \begin{subfigure}[b]{\textwidth}  % Adjust the width to fit your specific needs
        \centering
        \includegraphics[width=\textwidth]{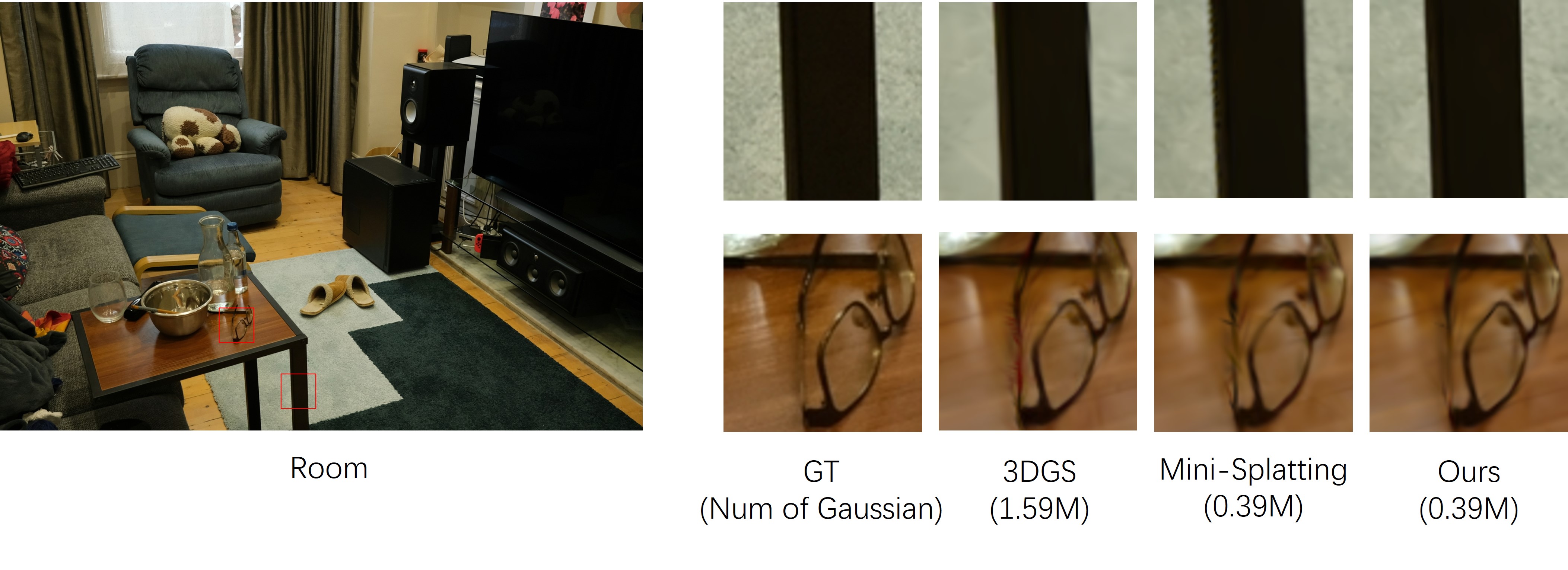}
        % \caption{Second subfigure}
        \label{fig:sub2}
    \end{subfigure}

    \begin{subfigure}[b]{\textwidth}  % Adjust the width to fit your specific needs
        \centering
        \includegraphics[width=\textwidth]{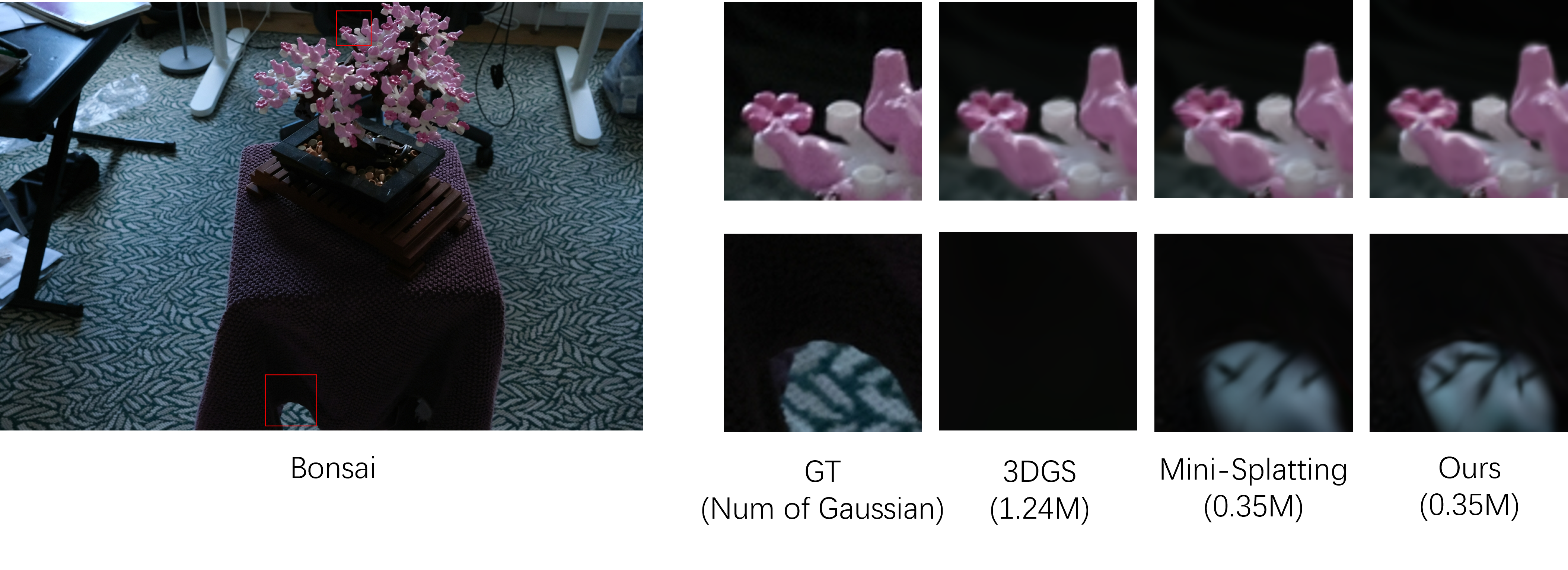}
        % \caption{Third subfigure}
        \label{fig:sub3}
    \end{subfigure}
    
    \caption{Visual comparison of three real-world scenes. Areas for zoomed-in comparison are marked with red boxes on the left images, highlighting the missing details in 3DGS and the poor representation of sharp features in Mini-Splatting.}
    \label{fig:three_subfigures}
\end{figure*}

% Details of the calculation are provided in the supplementary materials. \mlc{if we have space, put the formulation here.}

\textbf{Optimization}
In the original 3DGS framework \cite{kerbl20233d}, the limited representational power of individual Gaussians necessitates a strategy of increasing Gaussian density. This involves adaptively splitting or cloning Gaussians to better reconstruct complex regions. In contrast, our textured Gaussians demonstrate significantly enhanced representational power, enabling them to effectively capture local color and opacity variations. As a result, we have opted to eliminate the adaptive control of Gaussians in our study. This simplification streamlines the optimization process, making the optimization more straightforward and efficient.

In practice, we initiate our process using the output from Mini-Splatting \cite{fang2024mini} as a baseline, and then apply our Textured-GS method to enhance rendering quality. Mini-Splatting is chosen because it achieves results comparable to the original 3DGS while using fewer Gaussians. Additionally, it incorporates an intersection-preserving technique that discards Gaussians which do not directly intersect with the viewing ray. This method effectively minimizes scenarios where multiple low-opacity Gaussians overlap to blend a pixel, aligning well with our objective of maximizing the representational capability of each Gaussian.

We utilize a sigmoid activation function for both $\bm{\gamma}_c$ and $\bm{\gamma}_\alpha$ to ensure smooth gradients. The loss function is identical to the original 3DGS framework \cite{kerbl20233d}: 
\begin{equation}
    \mathcal{L} = (1 - \lambda) \mathcal{L}_1 + \lambda \mathcal{L}_{D-SSIM}
\end{equation}
where $\lambda$ is set to 0.2 in all experiments. 

The primary distinction in our gradient descent process, compared to the original framework's approach to view-dependent color, lies in the management of losses: In the original framework, view-dependent color loss from each processed pixel is aggregated and propagated to the SH coefficients in a single update at the end of each iteration. In contrast, our method updates the SH coefficients incrementally, applying the loss immediately after each pixel is processed. This modification is necessary because, unlike the view-dependent color approach where the input view vector remains constant for all pixels, the view vector in our spatially defined texture varies with each pixel. We apply this same incremental update mechanism to the opacity channel. Further details on the learning process are provided in the subsequent section.

\section{Experiments}

In this section, we evaluate our approach, detailing the implementation and discussing the results of ablation studies. The source code will be released upon the acceptance of the paper. 

\begin{figure*}
    \centering
    \includegraphics[width=1\linewidth]{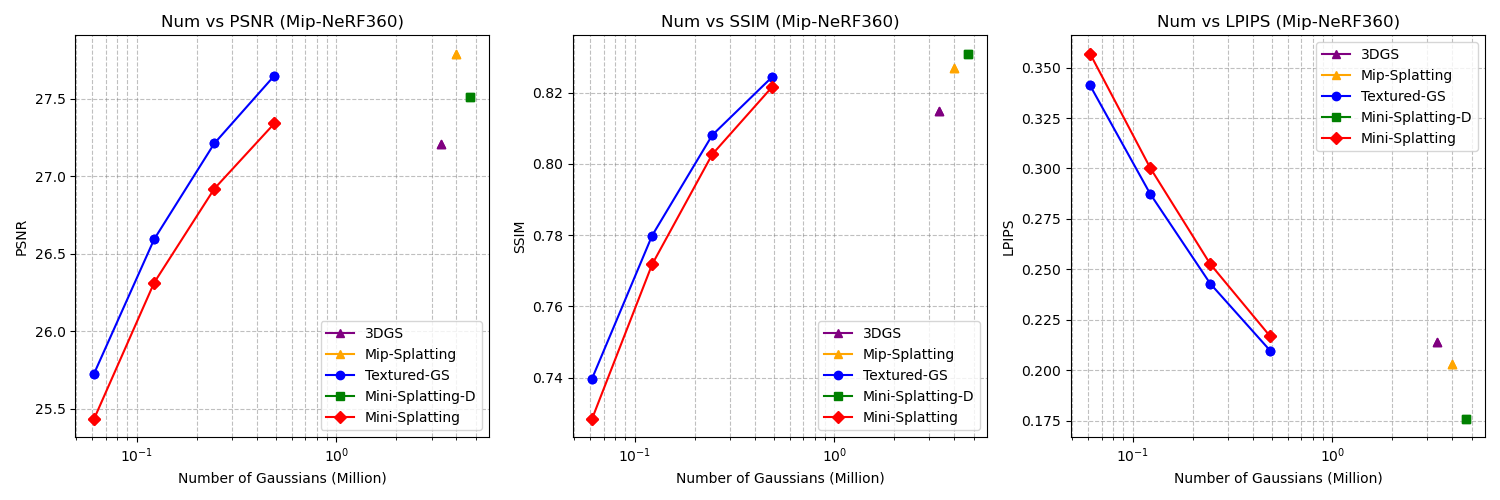}
    \caption{The rendering quality comparison between our method and Mini-Splatting across various numbers of Gaussians. Our method consistently outperforms Mini-Splatting across all scales and three distinct metrics.}
    \label{fig:NumMetric}
\end{figure*}

\subsection{Datasets}

We conducted experiments across three real-world datasets: Mip-NeRF360 \cite{barron2022mip}, Tanks and Temples \cite{knapitsch2017tanks}, and Deep Blending \cite{hedman2018deep}. To ensure consistency and fairness in our evaluations, we adhered to the same processing methodologies used in the 3DGS and Mini-Splatting studies, including identical scene selection, train/test splits, and image resolution settings. Specifically, for each dataset, we implemented a train/test division following the recommendations of Mip-NeRF 360 \cite{barron2022mip}, selecting every eighth photo for testing to facilitate consistent and meaningful comparisons. Our evaluations used standard error metrics widely used in the field, such as PSNR, LPIPS, and SSIM.

\subsection{Implementation Details}
We implemented Textured-GS using PyTorch and integrate it into the 3D Gaussian splatting (3DGS) \cite{kerbl20233d} optimization pipeline. As discussed previously, we removed the module for adaptive control of Gaussians. We also modified the Gaussian rasterization module within the 3DGS framework to facilitate the rendering of textured Gaussians and update SH coefficients through gradient descent.

Our experiments were carried out on an Ubuntu 20.04 platform with an NVIDIA RTX 4090 GPU. Starting with the output Gaussians from Mini-Splatting \cite{fang2024mini}, we trained our model for 14,000 iterations, applying a learning rate of 0.0025 for the color texture and 0.005 for the opacity texture, respectively. \zt{It is important to note that view dependent color was not loaded as part of our input due to a fundamentally different Gaussian setup.} We opted for the square root of the scale parameters, $\sqrt{s}$, for the ellipsoid in Equation \ref{eq:ellipsoid}, because larger ellipsoids tend to yield better performance, though this does not alter the actual Gaussian size in splatting. All other parameters were aligned with those used in Mini-Splatting and 3DGS to ensure a fair comparison.

\begin{table}
    \centering
    \resizebox{\linewidth}{!}{
    \begin{tabular}{|c|ccccc|}
    \hline
        Method & Num & FPS & Train Time & Memory & File Size \\
    \hline
        3DGS \cite{kerbl20233d} & 6.03M & 92 & 26m55s & 11.77GB & 1.5GB \\
        Mip-Splatting \cite{yu2024mip} & 7.88M & 62 & 40m43s & 15.21GB & 2.0GB \\
        Mini.-D \cite{fang2024mini} & 6.02M & 105 & 23m45s & 11.57GB & 1.5GB\\
        Mini. \cite{fang2024mini} & 0.53M & 416 & 11m33s & 5.46GB & 132MB\\
        Ours & 0.53M & 161 & 10m52s & 4.53GB & 164MB\\
    \hline
    \end{tabular}
    }
    \caption{Resource consumption of our method and baseline approaches on the Bicycle Scene from the Mip-NeRF 360 dataset. All the results are trained and inferred on Nvidia RTX 4090 GPU.}
    \label{tab:resource}
\end{table}

\begin{table*}
    \centering
    \begin{tabular}{|c|cc|ccc|}
    \hline
        Method & SH Degree for Color & SH Degree for Opacity &SSIM $\uparrow$ & PSNR $\uparrow$ & LPIPS $\downarrow$\\
    \hline
        Baseline & 0 & 0 & 0.810 & 26.78 & 0.224 \\
        Color Texture Only & 1 & 0 & 0.817 & 27.06 & 0.217\\
        Color Texture Only & 2 & 0 & 0.819 & 27.28 & 0.213\\
        Color Texture Only & 3 & 0 & 0.820 & 27.38 & 0.211 \\
        Color \& Opacity Texture & 3 & 1 & \colorbox{yellow}{0.822} & \colorbox{yellow}{27.44} & \colorbox{yellow}{0.211}\\
        Color \& Opacity Texture & 3 & 2 & \colorbox{orange}{0.823}  & \colorbox{orange}{27.57} & \colorbox{orange}{0.210}\\
        Color \& Opacity Texture (Ours) & 3 & 3 & \colorbox{red}{0.824} & \colorbox{red}{27.64} & \colorbox{red}{0.209}\\
    \hline
    \end{tabular}
    \caption{Ablation study assessing the impact of varying SH degrees on color and opacity using the Mip-NeRF 360 dataset. Employing SH degree zero for both components reduces the baseline to Mini-Splatting with monocolored Gaussians \cite{fang2024mini}. The highest SH degrees yield the best performance, clearly demonstrating the benefits of incorporating color and opacity textures. }
    \label{tab:Ablation}
\end{table*}

\subsection{Results}
\textbf{Rendering Quality} Table \ref{tab:All} presents a quantitative evaluation across three different real-world dataset. We compare our Textured-GS with the baseline method Mini-Splatting, 3DGS, and other related algorithms. In most categories, our Textured-GS outperforms both 3DGS and Mini-Splatting. The NeRF-based method Zip-NeRF \cite{barron2023zip}, currently the state-of-the-art on the Mip-NeRF 360 dataset, shows superior performance. Mini-Splatting-D \cite{fang2024mini}, a denser version of Mini-Splatting, and Mip-Splatting \cite{yu2024mip}, surpass our method in some categories but both use greater numbers of Gaussians than 3DGS, contradicting our goal of efficiency.

\begin{figure*}[t!]
    \centering
    \begin{subfigure}[b]{0.2\textwidth}
        \centering
        \includegraphics[width=1.0\textwidth]{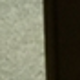}
        \caption{Ground Truth}
    \end{subfigure}
    \hfill
    \begin{subfigure}[b]{0.2\textwidth}
        \centering
        \includegraphics[width=1.0\textwidth]{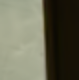}
        \caption{3DGS \cite{kerbl20233d}}
    \end{subfigure}
    \hfill
    \begin{subfigure}[b]{0.2\textwidth}
        \centering
        \includegraphics[width=1.0\textwidth]{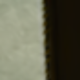}
        \caption{Mini-Splatting \cite{fang2024mini}}
    \end{subfigure}
    \hfill
    \begin{subfigure}[b]{0.2\textwidth}
        \centering
        \includegraphics[width=1.0\textwidth]{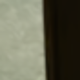}
        \caption{Ours}
    \end{subfigure}
    \caption{Zoomed-in comparison for the Room scene from the Mip-NeRF 360 dataset: Our proposed method, which incorporates textured opacity and color, achieves superior results in rendering sharp features with the same number of Gaussians compared to Mini-Splatting. Additionally, it more effectively captures the background texture than both 3DGS and Mini-Splatting.}
    \label{fig:Details}
\end{figure*}

Figure \ref{fig:three_subfigures} provides a visual comparison of three real-world scenes using our proposed method alongside two baseline methods. It shows that sharp object boundaries present significant challenges when represented with a limited number of Gaussians. For instance, in the Room scene, the table legs are effectively depicted using numerous Gaussians in the 3DGS method, but they appear serrated in Mini-Splatting due to a restricted number of Gaussians. Our method, which utilizes textured opacity across the Gaussian surface, effectively addresses this challenge, achieving a smoother representation with the same number of Gaussians, see Figure \ref{fig:Details} for more zoomed-in comparison. Additionally, 3DGS often completely misses detailed structures, as observed in the Train and Bonsai scenes, despite using a significantly larger number of Gaussians. 

Figure \ref{fig:NumMetric} further plots the performance of our approach compared to Mini-Splatting \cite{fang2024mini} on various numbers of Gaussians. Our method consistently outperforms the Mini-Splatting at each scale. Models with a fewer number of Gaussians are generated by using a lower sampling rate within the Mini-Splatting framework.

\textbf{Resource Consumption}
We assessed the resource consumption of various rendering methods, focusing on the number of Gaussians, frames per second (FPS), training time, and the maximum memory allocated during training. Table \ref{tab:resource} summarizes the comparative performance and efficiency of these methods. \zt{The results show that our approach uses less training memory and slightly more storage size than the baseline method with the same number of Gaussians. The reason is that no depth calculation is required in our framework, but four channels of SH coefficients are stored instead of three. It should be noted that both the rendering speed and the training time still have room for optimization.}

% The results show a clear reduction in memory usage. 

% It is worth noting that both the rendering speed and training time have significant room for optimization.

\subsection{Ablation Study}

We conducted an ablation study to evaluate the impact of two key components of our proposed method: textured color and textured opacity. For each component, we gradually increased the degree of Spherical Harmonics (SH) from zero to three.  Table \ref{tab:Ablation} summarizes the results, tested on the Mip-NeRF 360 dataset. Our method, employing the third degree level for both color and opacity, achieves the best results, demonstrating incremental improvements as the SH degree level increases.

\subsection{Limitations}
Our method enhances the capabilities of each Gaussian in the Mini-Splatting framework \cite{fang2024mini} by introducing textured color and opacity, leading to improved rendering quality. However, this approach has limitations, notably the increased computational cost due to additional calculations for ray-ellipsoid intersections, resulting in longer training and rendering times. We believe this limitation can be addressed by optimizing the rendering process and integrating the sampling for color and opacity into a single step, thereby streamlining computations and reducing overhead.

To isolate the impact of optimizing color and opacity textures, we kept the locations, sizes, and orientations of Gaussians as generated by Mini-Splatting unchanged. Additionally, we removed the adaptive control module for Gaussians, as found in 3DGS. Optimizing all these parameters in an end-to-end manner is likely to enhance the performance of our algorithm. As part of our future work, we plan to develop a comprehensive framework that starts with a structure-from-motion (SFM) point cloud, incorporates mechanisms for both densifying and simplifying Gaussians, and ultimately produces a set of textured Gaussians that optimally represent the scenes.

\section{Conclusion}
In conclusion, our study introduces Textured-GS, an innovative rendering method that leverages spherical harmonics to introduce spatially defined color and opacity variations within each Gaussian. This approach significantly enhances the visual quality of renderings, offering a more detailed and flexible solution compared to traditional Gaussian splatting methods. To validate the effectiveness of our method, we integrated Textured-GS into the Mini-Splatting framework without increasing the number of Gaussians. Our comprehensive evaluation across three real-world datasets demonstrates that Textured-GS consistently outperforms both the baseline Mini-Splatting and standard 3DGS in rendering quality. Additionally, we observed that our method efficiently addresses complex scenes that typically challenge Gaussian-based approaches, such as sharp edges and detailed structures. Moving forward, we plan to develop a fully optimized end-to-end framework that incorporates adaptive control of Gaussians, which we anticipate will further enhance the capabilities and efficiency of our approach, making it even more suitable for high-fidelity rendering tasks.

%%%%%%%%% REFERENCES
{\small
\bibliographystyle{ieee_fullname}
\bibliography{egbib}
}

\end{document}